\pdfoutput=1
\documentclass{article}

\usepackage{microtype}
\usepackage{graphicx}
\usepackage{subfigure}
\usepackage{booktabs} 

\usepackage{xcolor}

\usepackage{hyperref}

\usepackage[accepted]{icml2019}

\icmltitlerunning{Population Based Augmentation}

\begin{document}

\twocolumn[
\icmltitle{Population Based Augmentation: \\ Efficient Learning of Augmentation Policy Schedules}




\begin{icmlauthorlist}
\icmlauthor{Daniel Ho}{cal,googlex}
\icmlauthor{Eric Liang}{cal}
\icmlauthor{Ion Stoica}{cal}
\icmlauthor{Pieter Abbeel}{cal,cov}
\icmlauthor{Xi Chen}{cal,cov}
\end{icmlauthorlist}

\icmlaffiliation{cal}{EECS, UC Berkeley, Berkeley, California, USA}
\icmlaffiliation{cov}{covariant.ai, Berkeley, California, USA}
\icmlaffiliation{googlex}{Current affiliation: X, Mountain View, California, USA}

\icmlcorrespondingauthor{Daniel Ho}{daniel.ho@berkeley.edu}

\icmlkeywords{Machine Learning, ICML, PBA, Population Based Augmentation,  Population Based Training, augmentation, AutoAugment, neural network, CIFAR, SVHN, open source, ray, augmentation policy, augmentation schedule}

\vskip 0.3in
]



\printAffiliationsAndNotice{}  

\begin{abstract}
 A key challenge in leveraging data augmentation for neural network training is choosing an effective augmentation policy from a large search space of candidate operations. Properly chosen augmentation policies can lead to significant generalization improvements; however, state-of-the-art approaches such as AutoAugment are computationally infeasible to run for the ordinary user. In this paper, we introduce a new data augmentation algorithm, \emph{Population Based Augmentation} (PBA), which generates nonstationary augmentation policy schedules instead of a fixed augmentation policy.
We show that PBA can match the performance of AutoAugment on CIFAR-10, CIFAR-100, and SVHN, with three orders of magnitude less overall compute. On CIFAR-10 we achieve a mean test error of 1.46\%, which is a slight improvement upon the current state-of-the-art.
The code for PBA is open source and is available at \url{https://github.com/arcelien/pba}.

\end{abstract}

\section{Introduction}
\label{intro}

Data augmentation techniques such as cropping, translation, and horizontal flipping are commonly used to train large neural networks \cite{nin}. Augmentation transforms images to increase the diversity of image data. While deep neural networks can be trained on enormous numbers of data examples to exhibit excellent performance on tasks such as image classification, they contain a likewise enormous number of parameters, which causes overfitting. Data augmentation acts as a regularizer to combat this. However, most approaches used in training state-of-the-art networks only use basic types of augmentation. While neural network architectures have been investigated in depth \cite{alexnet, resnet, googlenet, vgg, wrn, densenet, pyramidnet}, less focus has been put into discovering strong types of data augmentation and data augmentation policies that capture data invariances.

A key consideration when applying data augmentation is picking a good set of augmentation functions, since redundant or overly aggressive augmentation can slow down training and introduce biases into the dataset \cite{Graham14a}. Many recent methods learn augmentation policies to apply different functions to image data. Among these, AutoAugment \cite{autoaug} stands out with state-of-the-art results in CIFAR-10 \cite{cifar}, CIFAR-100 \cite{cifar}, and ImageNet \cite{imagenet}. Using a method inspired by Neural Architecture Search \cite{NAS1}, Cubuk et al. learn a distilled list of augmentation functions and associated probability-magnitude values, resulting in a distribution of possible augmentations which can be applied to each batch of data. However, the search technique used in the work is very computationally expensive, and code has not been released to reproduce it. In this work, we address these issues with a simple and efficient algorithm for augmentation policy learning.

\begin{figure}[t]
  \centering
  \includegraphics[width=8.2cm]{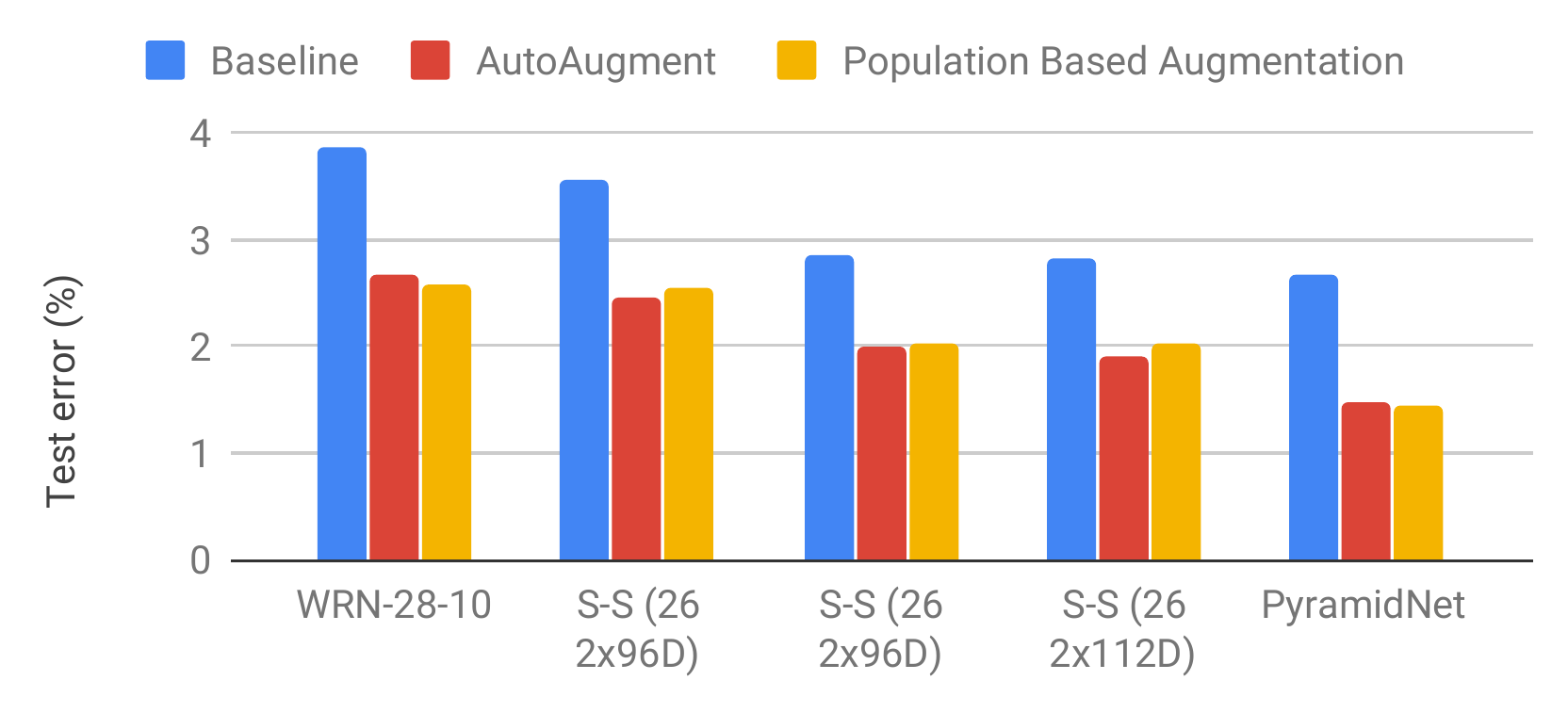}
     \vspace{-.3cm}
  \caption{PBA matches AutoAugment's classification accuracy across a range of different network models on the CIFAR-10 dataset, while requiring 1,000x less GPU hours to run. For the full set of results, refer to Table \ref{table-merged}. Assuming an hourly GPU cost of \$1.5, producing a new augmentation policy costs around \$7.5 for PBA vs \$7,500 with AutoAugment. The same scaling holds for the SVHN dataset as well.}
  \label{fig:chart}
\end{figure}

\begin{table}[t]
\caption{Comparison of pre-computation costs and test set error (\%) between this paper, AutoAugment (AA), and the previous best published results. Previous results did not pre-compute augmentation policies. AutoAugment reported estimated cost in Tesla P100 GPU hours, while PBA measured cost in Titan XP GPU hours. Besides PBA, all metrics are cited from \cite{autoaug}. For more detail, see Table \ref{table-merged}. *CIFAR-100 models are trained with the policies learned on CIFAR-10 data.}
\label{table-runtime}
\vskip 0.15in
\begin{center}
\begin{small}
\begin{tabular}{llccc}
\toprule
Dataset & Value & Previous Best & AA & PBA  \\
\midrule
CIFAR-10 & GPU Hours & - & 5000 & 5 \\
 & Test Error & 2.1 & 1.48 & 1.46 \\
\midrule
CIFAR-100 & GPU Hours & - & 0* & 0* \\
 & Test Error & 12.2 & 10.7 & 10.9 \\
\midrule
SVHN & GPU Hours & - & 1000 & 1 \\
 & Test Error & 1.3 & 1.0 & 1.1 \\
\bottomrule
\end{tabular}
\end{small}
\end{center}
\end{table}

Our formulation of data augmentation policy search, Population Based Augmentation (PBA), reaches similar levels of final performance on a variety of neural network models while utilizing orders of magnitude less compute. We learn a robust augmentation policy on CIFAR-10 data in five hours using one NVIDIA Titan XP GPU, and we visualize its performance in Figure \ref{fig:chart}. Relative to the several days it takes to train large CIFAR-10 networks to convergence, the cost of running PBA beforehand is marginal and significantly enhances results. These results are summarized in Table \ref{table-runtime}. PBA leverages the Population Based Training algorithm \cite{pbt} to generate an augmentation \textit{schedule} that defines the best augmentation policy for each epoch of training. This is in contrast to a fixed augmentation policy that applies the same transformations independent of the current epoch number.


We release code to run and evaluate our augmentation search algorithm at \url{https://github.com/arcelien/pba}. This allows an ordinary workstation user to easily experiment with the search algorithm and augmentation operations. A particularly interesting use case would be to introduce new augmentation operations, perhaps targeted towards a particular dataset or image modality, and be able to quickly produce a tailored, high performing augmentation schedule. Our code uses the Ray \cite{ray} implementation of PBT, which allows for easy parallelization across and within GPUs and CPUs.

This paper is organized as follows: First, we cover relevant background and AutoAugment (Section \ref{background}). We then introduce the PBA algorithm (Section \ref{augmentation}). We describe the augmentation schedules PBA discovers and its performance on several datasets. Finally, we seek to understand the efficiency gains of PBA through ablation studies and comparison with baseline methods (Section \ref{experiments}).

\section{Background}
\label{background}

\subsection{Related Work}
\label{related-work} 

We first review types of data augmentation for image recognition, which improve generalization with limited data by applying transformations to generate additional samples. Common techniques such as random cropping, flipping, rotating, scaling, and translating are used by top performing models for training on MINST, CIFAR-10, and ImageNet datasets \cite{aa_2, aa_30, aa_31, aa_32, alexnet, nin, pyramidnet}. Some additional approaches to generate augmented data include image combining \cite{samplepairing, manifold_4}, elastic distortions \cite{manifold_2}, and generative adversarial networks \cite{manifold_6}.

Augmentation has been shown to have a large impact on image modalities where data is scare or expensive to generate, like medical imaging \cite{manifold_8, manifold_9} or non-supervised learning approaches \cite{Mundhenk2018ImprovementsTC}.

Several papers have attempted to automate the generation of data augmentations with data-driven learning. These use methods such as manifold learning \cite{manifold}, Bayesian Optimization \cite{bayesian-data-aug}, and generative adversarial networks which generate transformation sequences \cite{ratner}. Additionally, \cite{smart-aug} uses a network to combine pairs of images to train a target network, and \cite{manifold_3} injects noise and interpolates images in an autoencoder learned feature space. AutoAugment \cite{autoaug} uses reinforcement learning to optimize for accuracy in a discrete search space of augmentation policies.

Our approach was inspired by work in hyperparameter optimization. There has been much previous work to well-tune hyperparameters, especially in Bayesian Optimization \cite{GP-UCB, TPE, Spearmint, SMAC}, which are sequential in nature and expensive computationally. Other methods incorporate parallelization or use non-bayesian techniques \cite{hyperband, vizier, shahg15, Springenberg2016BayesianOW, Gonzlez2016BatchBO} but still either require multiple rounds of optimization or large amounts of compute. These issues are resolved in Population Based Training \cite{pbt}, which builds upon both evolutionary strategies \cite{Clune2008NaturalSF} and random search \cite{Bergstra2012RandomSF} to generate non-stationary, adaptive hyperparameter schedules in a single round of model training.

\begin{figure}[t]
  \centering
  \includegraphics[width=8.2cm]{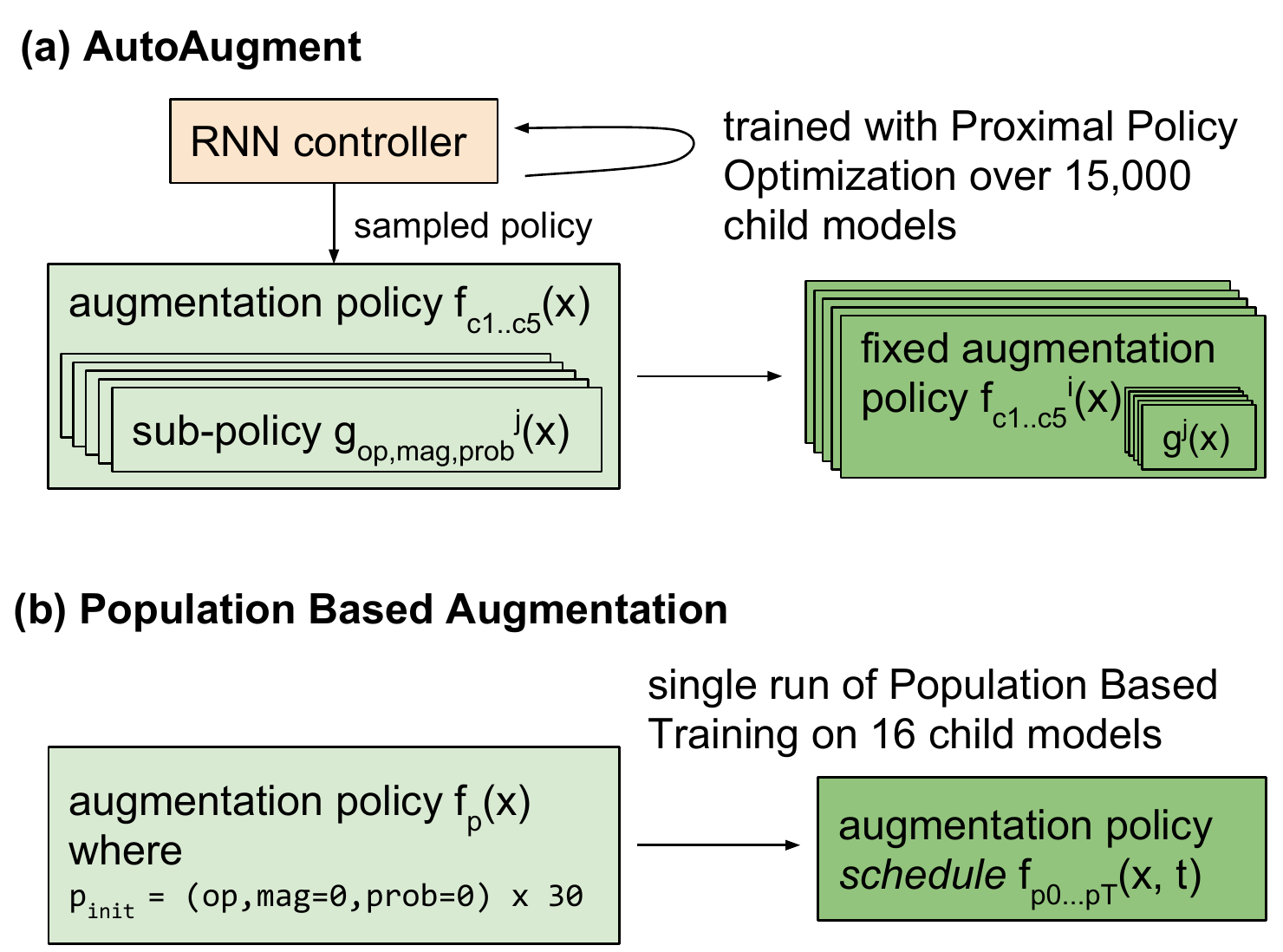}
  \caption{Comparison of AutoAugment and PBA augmentation strategies. In contrast to AutoAugment, PBA learns a schedule instead of a fixed policy. It does so in a short amount of time by using the PBT algorithm to jointly optimize augmentation policy parameters with the child model. PBA generates a single augmentation function $f(x, t)$ where $x$ is an input image and $t$ the current epoch, compared to AutoAugment's ensemble of augmentation policies $f^i(x)$, each of which has several further sub-policies.}
  \label{fig:autoaugment}
\end{figure}

\subsection{AutoAugment}
Cubuk et al. shows that using a diverse, stochastic mix of augmentation operations can significantly reduce generalization error. They automate the search over the space of data augmentation policies in a method called AutoAugment, which significantly improves neural network model accuracy on a variety of image datasets. AutoAugment follows an approach similar to work in the neural architecture search area \cite{nas2,enas} where a controller RNN network is trained via reinforcement learning to output augmentation policies maximizing for accuracy (Figure \ref{fig:autoaugment}). However, this approach is expensive in both time and compute, as the signal for the controller has to be generated by training thousands of models to convergence on different augmentation policies and evaluating final validation accuracy.

Cubuk et al. curated an augmentation policy search space based on operations from the PIL python library. These include ShearX/Y, TranslateX/Y, Rotate, AutoContrast, Invert, Equalize, Solarize, Posterize, Contrast, Color, Brightness, Sharpness, Cutout \cite{cutout}, and SamplePairing \cite{samplepairing}. Each operation has two associated parameters: probability and magnitude. The authors used discrete probability values from 0\% to 100\%, in increments of 10\%. Magnitude can range from 0 to 9 inclusive, but a few operations ignore this value and apply a constant effect. A policy would then consist of five sub-policies, each consisting of two operations and associated parameters. For every batch of data, one randomly selected sub-policy would be applied. In total, the final policy for AutoAugment concatenated the five best performing polices for a total of 25 sub-policies. 

To learn an augmentation policy, 15,000 sampled policies were evaluated on a Wide-ResNet-40-2 (40 layers, widening factor of 2) child model \cite{wrn} by taking the validation accuracy after training for 120 epochs on a ``reduced'' dataset. For CIFAR-10, this consists of 4,000 images from the training set, and for SVHN, 1,000 images. CIFAR-100 is trained with a transferred augmentation policy from CIFAR-10.

\begin{figure}[t]
  \centering
  \includegraphics[width=8.2cm]{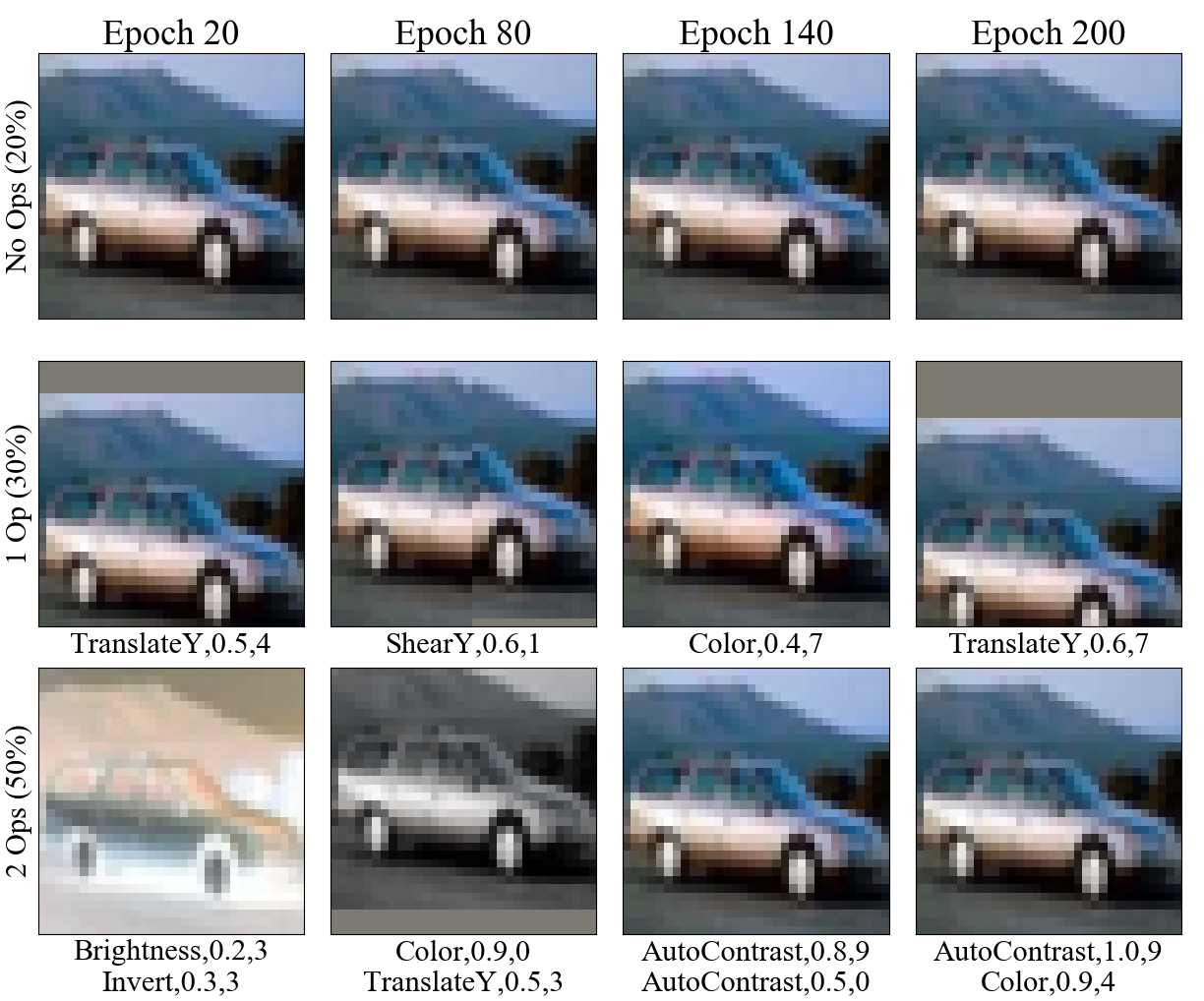}
  \caption{Augmentations applied to a CIFAR-10 ``car'' class image, at various points in our augmentation schedule learned on Reduced CIFAR-10 data. The maximum number of operations applied is sampled from 0 to 2. Each operation is formatted with name, probability, and magnitude value respectively.}
  \label{fig:aug-viz}
\end{figure}

\section{Population Based Augmentation}
\label{augmentation}

In this section we introduce the design and implementation of the PBA algorithm.

\subsection{Why Augmentation Schedules?}
The end goal of PBA is to learn a schedule of augmentation policies as opposed to a fixed policy. As we will see, this choice is responsible for much of the efficiency gains of PBA (Section \ref{experiments}). Though the search space for schedules over training epochs is larger than that of fixed policies $f \in F$ by a factor of $|F|^{|epochs|}$, counter-intuitively, PBA shows that it is far more efficient to search for a good schedule than a fixed policy. Several factors contribute to this.

First, estimating the final test error of a fixed augmentation policy is difficult without running the training of a child model to completion. This is true in particular because the choice of regularizing hyperparameters (e.g., data augmentation functions) primarily impacts the tail end of training. Therefore, estimating the final performance of a given fixed augmentation policy requires training a model almost to completion. In contrast, it is straightforward to reuse prior computations to estimate the performance of two variants of a schedule that share a prefix.

Second, there is reason to believe that it is easier to find a good augmentation policy when searching in the space of schedules. An augmentation function that can reduce generalization error at the end of training is not necessarily a good function at initial phases. Such functions would be selected out when holding the augmentation function fixed for the entirely of training. And though the space of schedules is large, most good schedules are necessarily \textit{smooth} and hence easily discoverable through evolutionary search algorithms such as PBT.

\subsection{Learning a Schedule}

In PBA we consider the augmentation policy search problem as a special case of hyperparameter schedule learning. Thus, we leverage Population Based Training (PBT) \cite{pbt}: a hyperparameter search algorithm which optimizes the parameters of a network \textit{jointly} with their hyperparameters to maximize performance. The output of PBT is not an optimal hyperparameter configuration but rather a trained model and schedule of hyperparameters. In PBA, we are only interested in the learned schedule and discard the child model result (similar to AutoAugment). This learned augmentation schedule can then be used to improve the training of different (i.e., larger and costlier to train) models on the same dataset.

PBT executes as follows. To start, a fixed population of models are randomly initialized and trained in parallel. At certain intervals, an ``exploit-and-explore'' procedure is applied to the worse performing population members, where the model clones the weights of a better performing model (i.e., exploitation) and then perturbs the hyperparameters of the cloned model to search in the hyperparameter space (i.e., exploration). Because the weights of the models are cloned and never reinitialized, the total computation required is the computation to train a single model times the population size.

The Ray framework \cite{ray} includes a parallelized implementation of PBT (\url{https://ray.readthedocs.io/en/latest/tune.html}) which handles the exploit-and-explore process in the backend. This implementation allows a user to deploy multiple trials on the same GPU, provided there is enough GPU memory. When the models only require a fraction of the computation resources and memory of an entire GPU, as in this work, training is sped up by fully utilizing the GPU.

\subsection{Policy Search Space}

\begin{algorithm}[tb]
  \caption{The PBA augmentation policy template, the parameters of which are optimized by PBT. The parameter vector is a vector of $(op, prob, mag)$ tuples. There are two instances of each $op$ in the vector, and this parameter cannot be changed. PBT learns a schedule for the $prob$ and $mag$ parameters during the course of training a population of child models.}
  \label{alg:apply}
\begin{algorithmic}
\vspace{.1cm}
  \STATE {\bfseries Input:} \textbf{data} $x$, \textbf{parameters}  $p$, [list of $(op, prob, mag)$]
  \STATE Shuffle parameters
  \STATE Set $count = $ [0, 1, 2] with probability [0.2, 0.3, 0.5]
  \FOR{$(op, prob, mag)$ in $p$}
    \IF{$count = 0$}
    \STATE break
    \ENDIF
  \IF{random(0, 1) $< prob$}
  \STATE $count = count - 1$
  \STATE $x = op(x, mag)$
  \ENDIF
  \ENDFOR
  \STATE \textbf{Return} $x$
\end{algorithmic}
\end{algorithm}

In Algorithm \ref{alg:apply}, we describe the augmentation policy function used in PBA and the optimization strategy we adapt from PBT. The challenge here is defining a smooth parameterization of the augmentation policy so that PBT can incrementally adopt good augmentations, while still allowing good coverage of the search space within a limited number of perturbations.

To make PBA more directly comparable with AutoAugment, we attempt to preserve the qualities of the AutoAugment formulation when possible, using the same augmentation functions, a similar number of total augmentation functions in the policy, and the same set of magnitude variants per function as applicable. Our augmentation policy search space consists of the augmentation operations from AutoAugment, less SamplePairing \cite{samplepairing}, for a total of 15 operations. We use the same code and magnitude options derived from PIL operations to ensure a fair comparison based on search algorithm performance. 

We define a set of hyperparameters consisting of two magnitude and probability values for each operation, with discrete possibilities for each. This gives us 30 operation-magnitude-probability tuples for a total of 60 hyperparameters. Like AutoAugment, we have 10 possibilities for magnitude and 11 possibilities for probability. When we apply augmentations to data, we first shuffle all operations and then apply operations in turn until a limit is reached. This limit can range from 0 to 2 operations.

Similar to the AutoAugment policy, PBA allows for two of the same augmentation operations to be applied to a single batch of data. Due to the use of a schedule, a single operation the PBA search space includes $(10 \times 11)^{30} \approx 1.75 \times 10^{61}$ possibilities, compared to $2.8 \times 10^{32}$ for AutoAugment. For discussion about the hyperparameter priors encoded within this policy template, see Section \ref{tuning}. Our policy template formulation is primarily motivated by the need to directly compare results with AutoAugment rather than optimizing for the best possible policy template.

\begin{algorithm}[tb]
  \caption{The PBA explore function. Probability parameters have possible values from 0\% to 100\% in increments of 10\%, and magnitdue parameters have values from 0 to 9 inclusive.}
  \label{alg:pbt-explore}
\begin{algorithmic}
  \STATE {\bfseries Input:} \textbf{Params} $p$, list of augmentation hyperparameters
  \FOR{$param$ in $p$}
  \IF{random(0, 1) $< 0.2$}
  \STATE Resample $param$ uniformly from domain
  \ELSE
  \STATE $amt = $ [0,1,2,3] uniformly at random
      \IF{random(0, 1) $< 0.5$}
      \STATE $param = param + amt$
      \ELSE
      \STATE $param = param - amt$
      \ENDIF
      \STATE Clip $param$ to stay in domain
  \ENDIF
  
  \ENDFOR
\end{algorithmic}
\end{algorithm}

\begin{table*}[t]
\caption{Test set error (\%) on CIFAR-10, CIFAR-100, and SVHN. Lower is better. The baseline applies regular random crop and horizontal flip operations. Cutout is applied on top of the baseline, and PBA/AutoAugment are applied on top of Cutout. We report the mean final test error of 5 random model initializations. We used the models: Wide-ResNet-28-10 \cite{wrn}, Shake-Shake (26 2x32d) \cite{shake-shake}, Shake-Shake (26 2x96d) \cite{shake-shake}, Shake-Shake (26 2x112d) \cite{shake-shake}, and PyramidNet with ShakeDrop \cite{pyramidnet, shake-drop}. Code for AA eval on SVHN was not released, so differences between our implementations could impact results. Thus, we report AA* from our re-evaluation.}
\label{table-merged}
\vskip 0.15in
\begin{center}
\begin{small}
\begin{tabular}{llccccc}
\toprule
Dataset & Model & Baseline & Cutout & AA & AA* & PBA  \\
\midrule
CIFAR-10 & Wide-ResNet-28-10     &   3.87 & 3.08 & 2.68 & & 2.58 $\pm$ 0.062  \\
 & Shake-Shake (26 2x32d) & 3.55 & 3.02 & 2.47&  & 2.54 $\pm$ 0.10  \\
 & Shake-Shake (26 2x96d) & 2.86 & 2.56 & 1.99& & 2.03 $\pm$ 0.11 \\
 & Shake-Shake (26 2x112d) & 2.82 & 2.57 & 1.89& & 2.03  $\pm$ 0.080  \\
 & PyramidNet+ShakeDrop   & 2.67 & 2.31 & 1.48& & \textbf{1.46  $\pm$ 0.077} \\
\midrule
Reduced CIFAR-10 & Wide-ResNet-28-10     &   18.84 & 17.05 & 14.13& & 12.82 $\pm$ 0.26  \\
& Shake-Shake (26 2x96d) & 17.05 & 13.40 & \textbf{10.04} & & 10.64 $\pm$ 0.22 \\
\midrule
CIFAR-100 & Wide-ResNet-28-10     &  18.8 & 18.41 & 17.09 & & 16.73 $\pm$ 0.15  \\
& Shake-Shake (26 2x96d) & 17.05 & 16.00 & 14.28 & & 15.31 $\pm$ 0.28 \\
& PyramidNet+ShakeDrop & 13.99 & 12.19 & \textbf{10.67} & & 10.94 $\pm$ 0.094 \\
\midrule
SVHN & Wide-ResNet-28-10 & 1.50 & 1.40 & 1.07 & 1.13 $\pm$ 0.024 & 1.18 $\pm$ 0.022 \\
& Shake-Shake (26 2x96d) & 1.40 & 1.20 & \textbf{1.02} & 1.10 $\pm$ 0.032 & 1.13 $\pm$ 0.029 \\
\midrule
Reduced SVHN & Wide-ResNet-28-10 & 13.21 & 32.5 & 8.15 & & 7.83 $\pm$ 0.22 \\
& Shake-Shake (26 2x96d) & 13.32 & 24.22 & \textbf{5.92} & & 6.46 $\pm$ 0.13 \\
\bottomrule
\end{tabular}
\end{small}
\end{center}
\vskip -0.1in
\end{table*}

\subsection{PBA Implementation}
We describe the formulation of our search in the format of PBT experiments \cite{pbt}.

\textbf{Step}: In each iteration we run an epoch of gradient descent.

\textbf{Eval}: We evaluate a trial on a validation set not used for PBT training and disjoint from the final test set.

\textbf{Ready}: A trial is ready to go through the exploit-and-explore process once 3 steps/epochs have elapsed.

\textbf{Exploit}: We use Truncation Selection \cite{pbt}, where a trial in the bottom 25\% of the population clones the weights and hyperparameters of a model in the top 25\%.

\textbf{Explore}: See Algorithm \ref{alg:pbt-explore} for the exploration function. For each hyperparameter, we either uniformly resample from all possible values or perturb the original value. 

\section{Experiments and Analysis}
\label{experiments}

In this section, we describe experiments we ran to better understand the performance and characteristics of the PBA algorithm. We seek to answer the following questions:

\begin{enumerate}
    \item How does classification accuracy and computational cost of PBA compare to state-of-the-art and random search baselines?
    \item Where does the performance gain of PBA come from -- does having a schedule of augmentations really matter, or is a stationary distribution sufficient?
    \item How does PBA performance scale with the amount of computation used?
    \item How sensitive is PBA to the hyperparameters of the optimization procedure -- did we just move part of the optimization process into hyperparameter selection?
\end{enumerate}

\subsection{Comparison with Baselines}
\textbf{Accuracy (CIFAR-10, CIFAR-100, SVHN)}
We first compare PBA to other state-of-the-art methods on the CIFAR-10 \cite{cifar} and SVHN \cite{svhn} datasets. Following \cite{autoaug}, we search over a ``reduced'' dataset of 4,000 and 1,000 training images for CIFAR-10 and SVHN respectively. Comparatively, CIFAR-10 has a total of 50,000 training images and SVHN has 73,257 training images with an additional 531,131 ``extra'' training images. PBA is run with 16 total trials on the Wide-ResNet-40-2 model to generate augmentation schedules.

For the augmentation policy, we initialize all magnitude and probability values to 0, as we hypothesize that less augmentation is required early in training when the validation accuracy is close to training accuracy. However, since training error decreases faster than validation error as training progresses, more regularization should be required, so we expect the probability and magnitude values to increase as training progresses. This would counteract overfitting as we introduce the model to more diverse data.

We then train models on the full training datasets, using the highest performing augmentation schedules discovered on the reduced datasets. The schedule learned on reduced CIFAR-10 is used to train final models on reduced CIFAR-10, CIFAR-10, and CIFAR-100. The schedule learned on reduced SVHN is used to train final models on reduced SVHN and SVHN. We report results in Table \ref{table-merged}. Each model is evaluated five times with different random initializations, and we report both the mean and standard deviation test set error in \%.

The models we evaluate on include: Wide-ResNet-28-10 \cite{wrn}, Shake-Shake (26 2x32d) \cite{shake-shake}, Shake-Shake (26 2x96d) \cite{shake-shake}, Shake-Shake (26 2x112d) \cite{shake-shake}, and PyramidNet with ShakeDrop \cite{pyramidnet, shake-drop}. PyramidNet with Shake-Drop uses a batch size of 64, and all other models use a batch size of 128. For Wide-ResNet-28-10 and Wide-ResNet-40-2 trained on SVHN, we use the step learning rate schedule proposed in \cite{cutout}, and for all others we use a cosine learning rate with one annealing cycle \cite{SGDR}. For all models, we use gradient clipping with magnitude 5. For specific learning rate and weight decay values, see the supplementary materials.

Additionally, we report Baseline, Cutout, and AutoAugment (AA) results found in \cite{autoaug}. For baseline, standard horizontal flipping and cropping augmentations were used. The training data is also normalized by the respective dataset statistics. For Cutout, a patch of size 16x16 is used for all CIFAR datasets, and size 20x20 for SVHN datasets. This applied with 100\% chance to each image. AutoAugment and PBA apply additional augmentations on top of the Cutout set (note that this possibly includes a second application of Cutout). The exception is Reduced SVHN, where the first 16x16 Cutout operation is removed as it was found to reduce performance.

\begin{figure}[t]
  \centering
  \begin{subfigure}[Operation magnitudes increase rapidly in the initial phase of training, eventually reaching a steady state around epoch 130.]{
  \includegraphics[width=8.2cm]{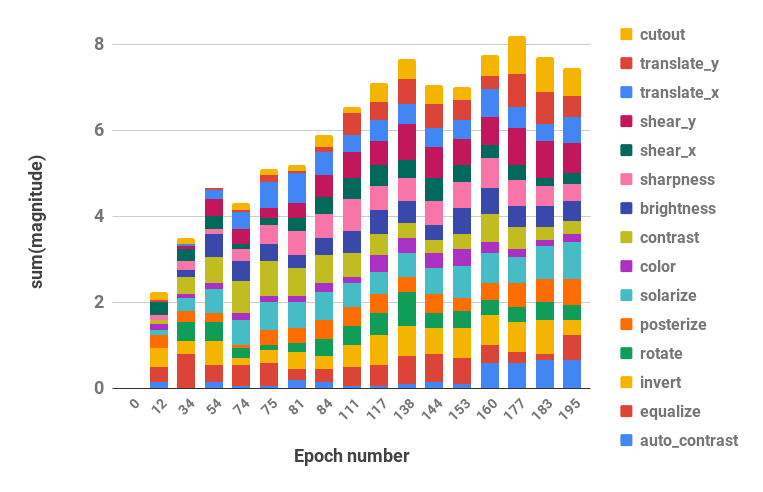}
  }
  \end{subfigure}
  \begin{subfigure}[Normalized plot of operation probability parameters over time. The distribution flattens out towards the end of training.]{
  \includegraphics[width=8.2cm]{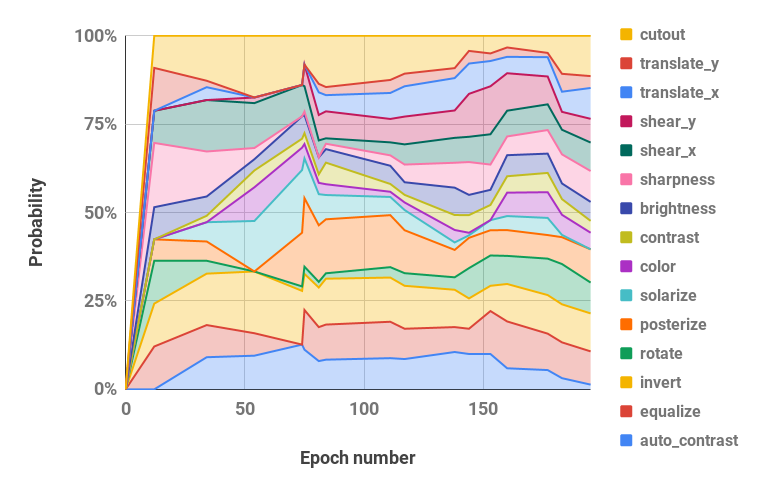}
  }
  \end{subfigure}
  \caption{Plots showing the evolution of PBA operation parameters in the discovered schedule for CIFAR-10. Note that each operation actually appears in the parameter list twice; we take the mean parameter value for each operation in this visualization.}
  \label{fig:pba-schedule}
\end{figure}

\textbf{CIFAR-10}
On Reduced CIFAR-10, we run PBA for 200 epochs, creating a policy schedule defined over 200 epochs. To extend the policy to Shake-Shake and PyramidNet models trained for 1800 epochs, we scale the length of the original schedule linearly.

While model accuracy on Reduced CIFAR-10 would have likely been improved with hyperparamater tuning for the reduced dataset size and smaller Wide-ResNet-40-2 model, our result shows that no hyperparameter tuning is required for high performance. 

Overall, the PBA learned schedule leads AutoAugment slightly on PyramidNet and Wide-ResNet-28-10, and performs comparably on Shake-Shake models, showing that the learned schedule is competitive with state-of-the-art.

We visualize the discovered schedule used in training our final CIFAR models in Figure \ref{fig:pba-schedule}. For the AutoContrast, Equalize, and Invert augmentations, magnitude values were ignored. From the probability values, our schedule seems to contain all augmentations to at least a moderate degree at some point, which is reasonable given our random perturb exploration method. However, there is emphasis on Cutout, Posterize, Invert, Equalize, and AutoContrast throughout the schedule. 

\cite{autoaug} suggests that color-based transformations are more useful on CIFAR compared to geometric ones, and our results also indicate this. However, they also found that the Invert transformation is almost never used, while it was very common in our schedule. A possible explanation may be that a model is able to better adapt to Invert when using a nonstationary policy. PBA may be exploring systematically different parts of the design space than AutoAugment. Alternatively, it may be that by the randomness in PBA, Cutout was introduced and impacted performance. It may be fruitful to explore combinations of PBA and AutoAugment to design nonstationary policies with more precision from a RNN Controller.

\textbf{CIFAR-100}
We additionally evaluate on CIFAR-100 using the same augmentation schedule discovered using Reduced CIFAR-10. We find that these results are also competitive with AutoAugment and significantly better than Baseline or only applying Cutout.

\begin{figure}[t]
  \centering
  \includegraphics[width=8.2cm]{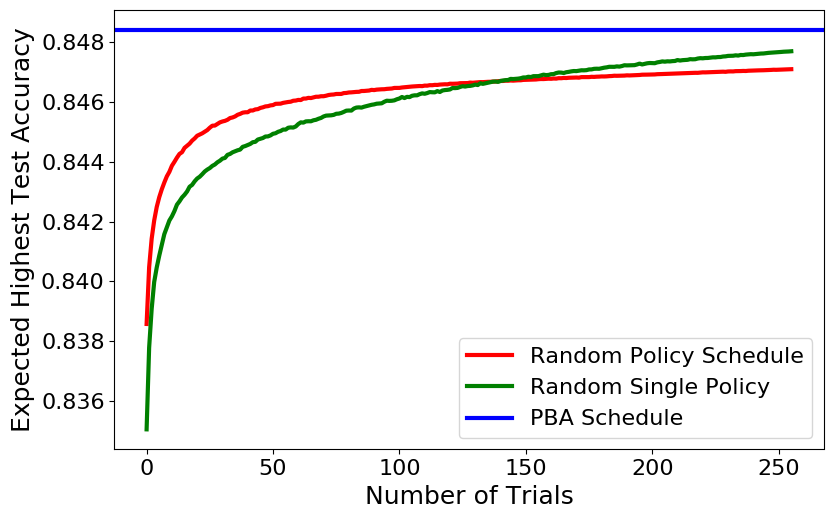}
  \caption{Plot of the expected best child test accuracy after a given number of random trials on Wide-ResNet-40-2. Random policy schedules were generated by randomly selecting intervals of length between 1 and 40, and then selecting a random policy for the interval. All values were selected uniformly from the domain.}
  \label{fig:random-search}
\end{figure}

\textbf{SVHN}
We ran PBA for 160 epochs on a 1,000 image Reduced SVHN dataset to discover an augmentation policy schedule without tuning any parameters of the algorithm. See the appendix for a visualization of an example PBA policy on the SVHN dataset. 

We then trained models on both the Reduced SVHN and SVHN Full (core training data with extra data), using the discovered schedule. Except for the Wide-ResNet-28-10 model on Reduced SVHN, training was done without tuning, using the hyperparamters from AutoAugment. We were able to obtain a policy comparable with AutoAugment. This demonstrates the robustness of the PBA algorithm across datasets.

Examining the learned policy schedule, we observe that Cutout, Translate Y, Shear X, and Invert stand out as being present with high probability across all epochs. This fits with the findings of \cite{autoaug}, indicating that Invert and geometric transformations are successful in SVHN because it is important to learn invariances to these augmentations. From another perspective, all of the augmentations appear with reasonable probability at some point in the schedule, which suggests that using a preliminary strategy like AutoAugment to filter out poor performing augmentations would be an interesting direction to explore.

\begin{table*}[ht]
\caption{Ablation study: We evaluate models on CIFAR-10 using a fixed policy (the last policy of the PBA schedule learned on Reduced CIFAR-10), shuffled schedule order, and a fully collapsed schedule, comparing to results with the original PBA schedule. See Section \ref{sec-schedule-matter} for further explanation. We evaluate each model once, and some combinations were not evaluated due to cost considerations.}
\label{table-cifar10-ablation-fixed}
\vskip 0.15in
\begin{center}
\begin{small}
\begin{tabular}{lcccccc}
\toprule
Model & Cutout & Fixed Policy & Order-shuffled & Fully-shuffled & PBA \\
\midrule
Wide-ResNet-28-10     & 3.08 &  2.76  & 2.66 & 2.89 &  2.576 $\pm$ 0.062  \\
Shake-Shake (26 2x32d) & 3.02 &2.73 & -&- & 2.54 $\pm$ 0.10  \\
Shake-Shake (26 2x96d) & 2.56 & 2.33 &- &- & 2.03 $\pm$ 0.11 \\
Shake-Shake (26 2x112d) & 2.57 &2.09 &- &- & 2.03  $\pm$ 0.080  \\
PyramidNet+ShakeDrop   & 2.31 & 1.55 & - &- & 1.46  $\pm$ 0.077 \\
\bottomrule
\end{tabular}
\end{small}
\end{center}
\vskip -0.1in
\label{table:ablate_schedule}
\end{table*}

\textbf{Computational Cost} AutoAugment samples and evaluates $\sim$15,000 distinct augmentation policies on child models, which requires about $15000 * 120 = 1.8m$ epochs of training. In comparison, PBA leverages PBT to learn a schedule with a population of 16 child models. PBA uses 200 epochs of training per child model, for a total of $3200$ epochs, or over 500x less compute than AutoAugment.

As a second baseline, we also train 250 child models with randomly selected augmentation policies, and 250 child models with randomly selected augmentation schedules. In Figure \ref{fig:random-search}, we use this data to plot the expected maximum \textit{child model} test accuracy after a given number of random trials. As shown, it takes over 250 trials for the expected child accuracy, which is strongly correlated with final accuracy, to approach that reached by a single 16-trial PBA run. Hence, PBA still provides over an order of magnitude speedup here.


\textbf{Real-time Overhead} Since PBT trains all members of its population simultaneously, the minimal real-time overhead is just the time it takes to train one child model. In practice, there is a slight overhead from the mutation procedures triggered by PBT, but the overall search time is still small compared to the time to train the primary model. In contrast, AutoAugment leverages reinforcement-learning based techniques, in which a Recurrent Neural Network (RNN) controller is trained with the reinforcement learning algorithm Proximal Policy Optimization (PPO) \cite{ppo}. Using this strategy, new augmentation policies can only be sampled and trained after the previous batch of samples has completed, so parallelization is limited to the batch size of the PPO update.

\subsection{Does having a schedule matter?}
\label{sec-schedule-matter}

PBA distinguishes itself from AutoAugment by learning a augmentation policy \textit{schedule}, where the distribution of augmentation functions can vary as a function of the training epoch. To check whether a schedule contributes to performance, we try training the model using (1) the last augmentation policy of the PBA schedule as a fixed policy, (2) the augmentation schedule with the order of policies shuffled but the duration of each policy fixed, and (3) the augmentation schedule collapsed into a time-independent stationary distribution of augmentations (i.e., a policy is sampled independently for each batch of data, where each policy is weighted by its duration).

In Table \ref{table:ablate_schedule}, we see that training with the PBA Fixed Policy degrades accuracy by $\sim$10\% percent on average, which is significantly worse than training with the full schedule. Compared to using Cutout, the fixed policy gives up $\sim$50\% of gains on Wide-ResNet-28-10, Shake-Shake 32, and Shake-Shake 96, and $\sim$10\% of gains on Shake-Shake 112 and PyramidNet. This shows that the augmentation schedule improves accuracy over a fixed policy, especially on smaller models.

Similarly, when we evaluated the shuffled schedules (only on Wide-ResNet-28-10), accuracy is also significantly lower, showing that a stationary distribution derived from the schedule does not emulate the schedule. We hypothesize that schedule improves training by allowing "easy" augmentations in the initial phase of training while still allowing "harder" augmentations to be added later on.

\subsection{Hyperparameter Tuning and Sensitivity}
\label{tuning}
We did not tune the discrete space for magnitude or probability options to keep our policy easy to compare to AutoAugment. We have two copies of each operation, as the AutoAugment sub-policy is able to contain two copies of the same operation as well.

For the search algorithm, we lightly tuned the explore function and the distribution for $count$ in Algorithm \ref{alg:apply}, the maximum number of augmentation functions to apply for each batch of data. While we keep the maximum value of $count$ at 2 in line with AutoAugment's length 2 subpolicy, there may be room for performance improvement by carefully tuning the distribution.

We tried perturbation intervals of 2 and 4 once, but did not find this value to be sensitive. We also tried to run PBT for 100 epochs, but found this to slightly decrease performance when evaluated on models for 200 epochs. 

It may be interesting to consider training a larger child model (e.g, Shake-Shake) for 1,800 epochs to generate a schedule over the full training duration and eliminate the need to stretch the schedule. In a similar vein, an experiment to use PBT directly on the full CIFAR-10 dataset or Wide-ResNet-28-10 model may lead to better performance, and is computationally feasible with PBA.

\section{Conclusion}

This paper introduces PBA, a novel formulation of data augmentation search which quickly and efficiently learns state-of-the-art augmentation policy schedules. PBA is simple to implement within any PBT framework, and we release the code for PBA as open source.

\section*{Acknowledgements}
We thank Richard Liaw, Dogus Cubuk, Quoc Le, and the ICML reviewers for helpful discussion.



\nocite{pbt}

\bibliography{ms}
\bibliographystyle{icml2019}

\clearpage

\appendix
\section{PBA Scalability with Compute}
\label{section-compute}

\begin{table*}[t]
\caption{Test error during PBT search and policy schedule evaluated afterwards, for varying population sizes and models. PBA Search with variation of model and compute, on Reduced CIFAR-10 dataset. ResNet-20 (Res) took approximately half the compute of WideResNet-40-2 (WRN). Number in title is the population size, and speedup is relative to AutoAugment. Note that models with larger population sizes, while scoring high during the search, don't actually perform better when re-evaluated.}
\label{table-cifar10-ablation-comp}
\vskip 0.15in
\begin{center}
\begin{small}
\begin{tabular}{lcccccr}
\toprule
Model & 8-Res & 16-Res & 32-Res & 16-WRN & 32-WRN & 64-WRN  \\
\midrule
WRN-40-2 during search & - & - & - & 0.8484 & 0.8446 & 0.8523  \\
WRN-40-2 & - & - & - & 0.8452 & 0.8445 & 0.8446   \\
ResNet-20 during search & 0.7484 & 0.7657 & 0.7619 & - & - & -  \\
ResNet-20 & 0.7457 & 0.7545 & 0.7534 & - & - & -    \\
WRN-28-10 & 0.9711 & 0.9721 & 0.9740 & 0.9740 & 0.9736 & 0.9703 \\
\midrule
Relative Speedup & 2250x & 1125x & 562.5x & 562.5x & 281.25x & 140.625x \\
\bottomrule
\end{tabular}
\end{small}
\end{center}
\vskip -0.1in
\end{table*}

\begin{figure}[t]
  \centering
  \includegraphics[width=8.2cm]{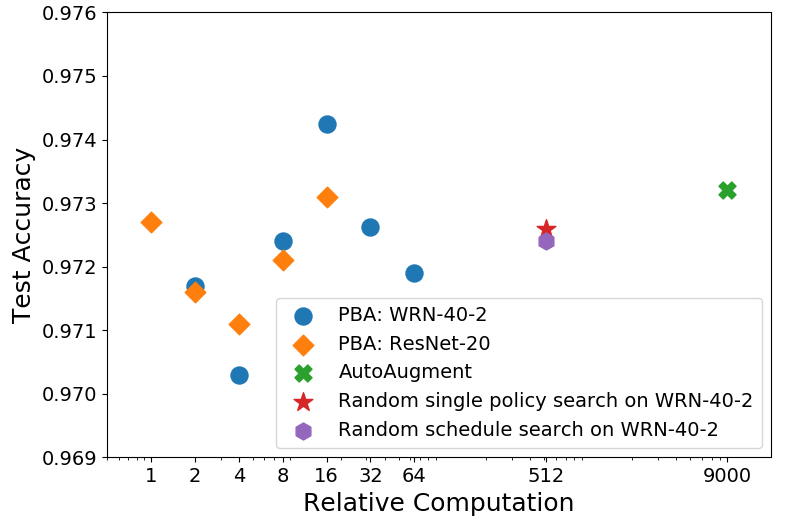}
  \caption{Comparison of relative cost of search to test accuracy of augmentation policies evaluated on WideResNet-28-10. Child model WRN-40-2 was evaluated for population sizes from 2 to 64, and ResNet-20 was evaluated for sizes 2 to 32. All policies were trained on Reduced CIFAR-10.}
  \label{fig:compute}
\end{figure}

\begin{figure}[t]
  \centering
  \includegraphics[width=8.2cm]{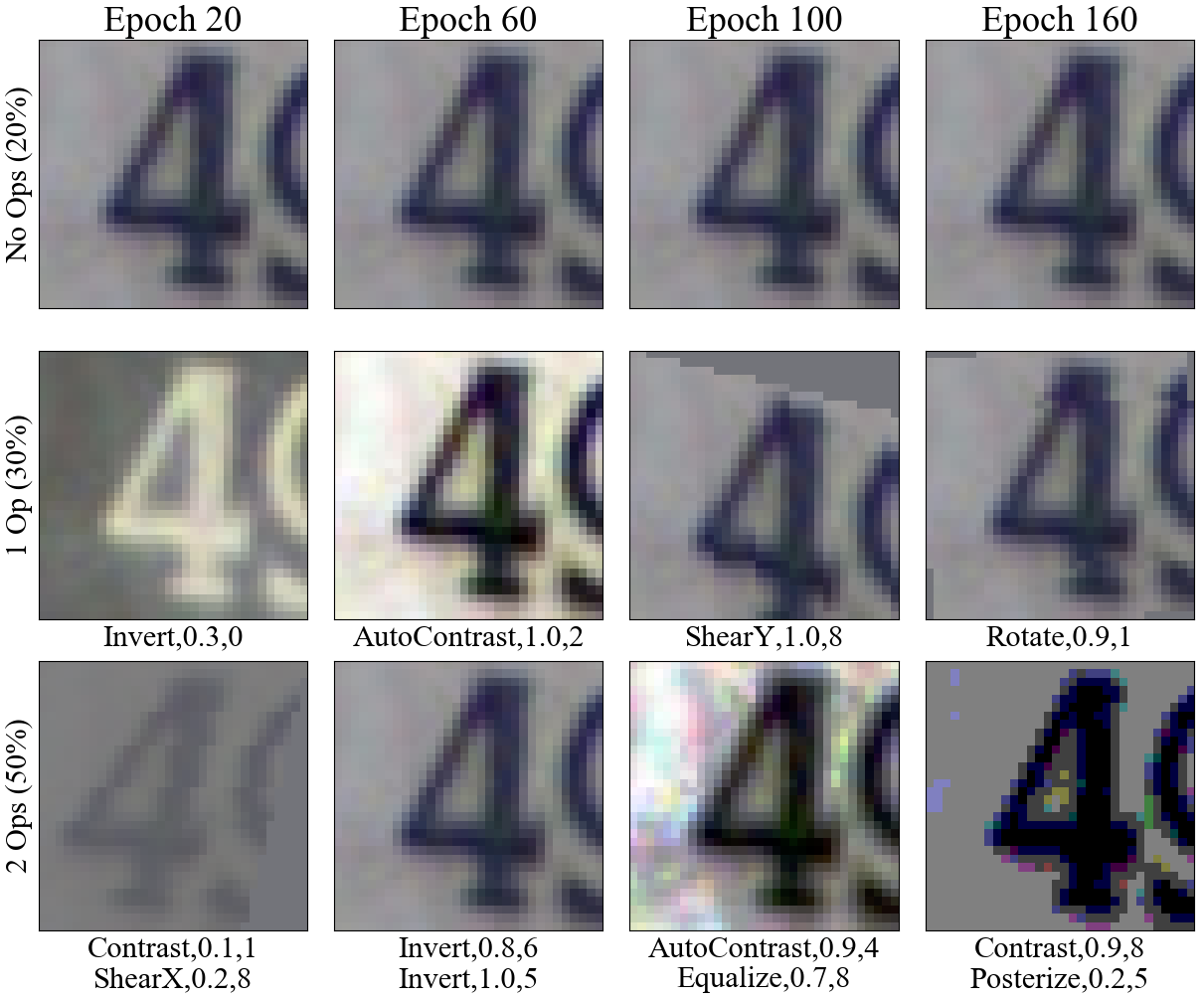}
  \caption{Augmentations applied to a SVHN ``4'' class image, at various points in our augmentation schedule learned on Reduced SVHN data. Each operation is formatted with name, probability, and magnitude value respectively.} 
  \label{fig:svhn-viz}
\end{figure}

\begin{table*}[t]
\caption{Hyperparameters used for evaluation on CIFAR-10, CIFAR-100, and (R)educed-CIFAR-10. Besides Wide-ResNet-28-10 and Wide-ResNet-40-2 on Reduced SVHN, no hyperparameter tuning was done. Instead, all hyperparameters are the same as those used in AutoAugment.}
\label{table-hp-cifar10}
\vskip 0.15in
\begin{center}
\begin{small}
\begin{tabular}{ccccc}
\toprule
Dataset & Model & Learning Rate & Weight Decay & Batch Size \\
\midrule
CIFAR-10 & Wide-ResNet-40-2    &   0.1 & 0.0005 & 128 \\
CIFAR-10 & Wide-ResNet-28-10    &   0.1 & 0.0005 & 128 \\
CIFAR-10 & Shake-Shake (26 2x32d) & 0.01 & 0.001 & 128 \\
CIFAR-10 & Shake-Shake (26 2x96d) & 0.01 & 0.001 & 128 \\
CIFAR-10 & Shake-Shake (26 2x112d) & 0.01 & 0.001 & 128 \\
CIFAR-10 & PyramidNet+ShakeDrop   & 0.05 & 0.00005 & 64 \\
CIFAR-100 & Wide-ResNet-28-10    &   0.1 & 0.0005 & 128 \\
CIFAR-100 & Shake-Shake (26 2x96d) & 0.01 & 0.0025 & 128 \\
CIFAR-100 & PyramidNet+ShakeDrop   & 0.025 & 0.0005 & 64 \\
R-CIFAR-10 & Wide-ResNet-28-10    &   0.05 & 0.005 & 128 \\
R-CIFAR-10 & Shake-Shake (26 2x96d) & 0.025 & 0.0025 & 128 \\
SVHN & Wide-ResNet-40-2    &   0.05 & 0.005 & 128 \\
SVHN & Wide-ResNet-28-10 & 0.005 & 0.001 & 128 \\
SVHN & Shake-Shake (26 2x96d) & 0.01 & 0.00015 & 128 \\
R-SVHN & Wide-ResNet-28-10 & 0.05 & 0.01 & 128 \\
R-SVHN & Shake-Shake (26 2x96d) & 0.025 & 0.005 & 128 \\
\bottomrule
\end{tabular}
\end{small}
\end{center}
\vskip -0.1in
\end{table*}

\begin{figure}[t]
  \centering
  \begin{subfigure}[Operation magnitudes.]{
  \includegraphics[width=8.2cm]{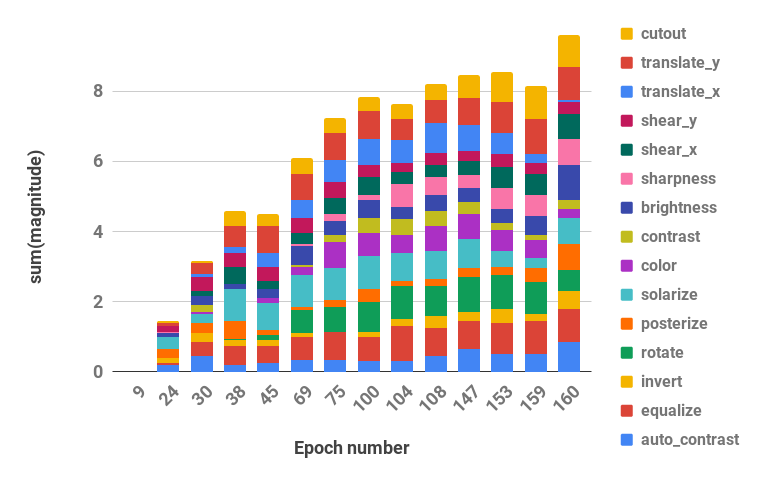}
  }
  \end{subfigure}
  \begin{subfigure}[Normalized plot of operation probability parameters over time.]{
  \includegraphics[width=8.2cm]{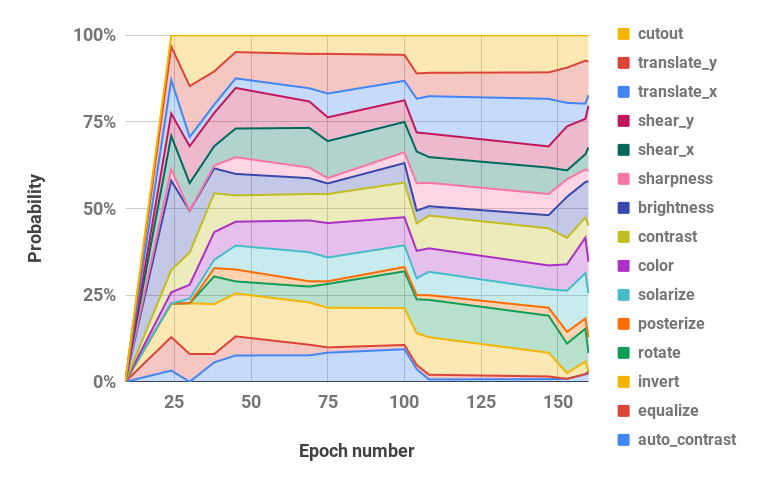}
  }
  \end{subfigure}
  \caption{Plots showing the evolution of PBA operation parameters in a schedule learned on Reduced SVHN. Note that each operation actually appears in the parameter list twice; we take the mean parameter value for each operation in this visualization.}
  \label{fig:pba-svhn-schedule}
\end{figure}

In Figure \ref{fig:compute} we look at how large the model and PBT population size is necessary to learn an effective schedule. The population size determines how much of the search space is explored during training, and also the computational overhead of PBA. Our results indicate that a population size of 16 WRN-40-2 models performs the best. Having more than 16 trials seems not to help, and having less than 16 seems to lead to decreased performance. However, we found that results could fluctuate significantly between runs of PBT, most likely due to exploring a very limited search space with a noisy exploration strategy. 

Besides WRN-40-2, we also tried to use a ResNet-20 \cite{resnet-v2} model for PBT population, which required about half the compute. Empirical results (in  Table \ref{table-cifar10-ablation-comp} and Figure \ref{fig:compute}) suggest that the ResNet-20 population does not achieve as high of a test accuracy as with WRN-40-2, but results were relatively close. Because a ResNet-20 model has much less parameters, training accuracy plateaus faster than WRN-40-2, which may change the effects of augmentation.

\section{Model Hyperparameters}
The hyperparameters used to train WideResNet-40-2 to discover augmentation schedules, and also the ones used to train final models, are displayed in Table \ref{table-hp-cifar10}. For full details on the hyperparameters and implementation, see the open source code.

\section{SVHN discovered schedule}
See Figure \ref{fig:svhn-viz} for a visualization of the policy on an example image and Figure \ref{fig:pba-svhn-schedule} for a visualization of an example PBA policy on the SVHN dataset.

Examining the learned policy schedule, we observe that Cutout, Translate Y, Shear X, and Invert stand out as being present with high probability across all epochs. This fits with the findings of \cite{autoaug} indicating that Invert and geometric transformations are successful in SVHN because it is important to learn invariances to these augmentations. From another perspective, all of the augmentations appear with reasonable probability at some point in the schedule, which suggests that using a preliminary strategy like AutoAugment to filter out poor performing augmentations would be an interesting direction to explore.

\end{document}